\documentstyle[eaclap]{article}

\title{\vspace{-0.5in}An Algorithm to Co-Ordinate Anaphora Resolution and PPS
Disambiguation Process}

\author{Saliha Azzam\\
CRIL Ingénierie -  CAMS La Sorbonne \\
174, Rue de la République\\
92817 Puteaux\\
France\\
azzam@cril-ing.fr\\ }

\begin{document}

\maketitle
\vspace{-0.5in}
\begin{abstract}

Both anaphora resolution and prepositional phrase (PP) attachment are
the most frequent ambiguities in natural language processing.  Several
methods have been proposed to deal with each phenomenon separately,
however none of proposed systems has considered the way of dealing
both phenomena We tackle this issue here, proposing an algorithm to
co-ordinate the treatment of these two problems efficiently, i.e., the
aim is also to exploit at each step all the results that each
component can provide.

\end{abstract}

\section{Introduction}

Several methods have been proposed to deal with anaphora resolution
and prepositional phrase (PP) attachment phenomenon and separately, so
that the literature is very abundant : for PPs see e.g., (Frazier and
Fodor, 1979; Hobbs, 1990; Wilks and Huang, 1985), and for
anaphora see e.g., (Carter, 1986; Reinhart, 1983; Sidner,
1983). However none of these methods has considered the way of dealing
both phenomena in the same concrete system.

We propose in this paper an algorithm that deals with both phenomena,
in the same analyser.  The anaphora module pertains to the recent
methods, uses a set of resolution rules based on the focusing
approach, see (Sidner, 1983). These rules are applied to the
conceptual representation and their output is a set of candidate
antecedents. Concerning the PPs, unattached prepositions involve empty
or unfilled roles in the Conceptual Structures (CSs), expressed in a
frame-based language (Zarri, 1992). The disambiguation procedure aims
at filling the empty roles using attachment rules.

This work was accomplished in the context of COBALT project (LRE 61-011
), dealing with financial news. A detailed discussion about both
procedures of anaphora resolution and PP attachment is largely
developed in (Azzam, 1994).

\section{The algorithm}

Two of the main principles of the algorithm are :

\bigskip

\noindent
a) The algorithm is applied on the text sentence by sentence, i.e. the
ambiguities of the previous sentences have already been considered
(resolved or not).

\bigskip

\noindent
b) The anaphora procedure skips the resolution of a given anaphor when
this anaphor is preceded by an unattached preposition. This is because
the resolution rules may have an empty role as a parameter, due to
this unattached preposition. The resolution of the anaphor is then
postponed to the second phase of anaphora resolution.

The proposed procedure is based on successive calls to the anaphora
module and to the PP attachment module.  The output of each call is a
set of CSs that represent the intermediate results exchanged between
each call and on which both modules operate in turn. The aim is to
fill the unfilled roles in the CSs, due to anaphora or unattached
PPs. To summarize the algorithm is:

\bigskip
\begin{itemize}
\item[] 1) Apply the anaphora module first.

\item[] 2) Apply the PP attachment procedure.

\item[] 3) If some anaphora are left unresolved, apply the anaphora
  module again.

\item[] 4) If there are still unattached PPs, apply the attachment
  procedure again.

\item[] 5) Repeat (3) and (4), until all PPs and anaphors are treated.
\end{itemize}

\bigskip

The order in which the two modules are called is based on efficiency
deduced from statistical data performed on  COBALT corpuses.

\bigskip

Three main cases are faced by the algorithm :

\bigskip

\noindent
a) When the anaphor occurs before a given preposition in the sentence,
its resolution does not depend on where the preposition is to be
attached (except for cataphors that are quite rare). In this case the
anaphora module can be applied before the attachment procedure.

The example 1 below shows that the resolution of the anaphoric pronoun
that must be performed first and that the PP starting with of be
attached later.

(1) {\it The sale of Credito was first proposed last August and
\underline{that} \underline{of} BCI
late last year.}

\bigskip

\noindent
b) When the anaphor occurs after one or several unattached
prepositions, it could be an intra-sentential anaphor (i.e. referring
to an entity in the same sentence), then its resolution may depend on
one of the previous prepositional phrases. In this case, the
resolution of the anaphora is postponed to a next call of the anaphora
module according to principle b) stated above.

\bigskip

\noindent
c) When the anaphor is included in a PP (particular case of b), PP
attachment rules need semantic information about the ``object'' of the
PP; when it is a pronoun, no semantic information is available, so
that the attachment rules can not be applied. The anaphoric pronouns
have to be resolved first, so as to determine what semantic class they
refer to ; the PP attachment procedure can then be applied. When a
sequence contains more than two such PPs, i.e., with anaphors as
objects, the length of a cycle is more than 4.

\section{An example}

{\it (2) UPHB shares have been suspended \underline{since} October 29
\underline{at} the
  firm's request following a surge \underline{in} \underline{its} share price
\underline{on} a takeover
  rumour.}

\bigskip

\noindent
- The pronoun \underline{its} can not be resolved by the anaphora resolution
module because it is preceded by unattached PPs ; its resolution is
skipped.

\bigskip

\noindent
- The PP attachment procedure is then called to determine the
attachment of \underline{since} and \underline{at} while the object of the
\underline{in} PP comprises an
anaphoric pronoun \underline{its} (case c) and the \underline{on} PP is
preceded by \underline{its}. The attachment of both PPs is then skipped.

\bigskip

\noindent
- The anaphora module is called again to resolve the anaphoric pronoun
\underline{its}, which is possible, in this example, since the previous PPs
have
been attached and there is no anaphors before.

\bigskip

\noindent
- Finally, the PP attachment procedure has to be called again for the
\underline{in} and \underline {on} PPs.

\bigskip

Notice that even if each module is called several times, there
is no redundancy in the processing. The algorithm should be considered
as the splitting of both anaphora resolution and PP attachment
procedures into several phases and not as the repetition of each
procedure.

\section{Conclusion}

The objective was to emphasise more than it has been done until now,
the fact that PP attachment and anaphora resolution could interact in
the same system in order to produce a complete conceptual analysis,
instead of slowing down each other. The algorithm we proposed in this
paper, is independent of the used approaches in both anaphora and
attachment modules. It concerns rather the way of managing the
interaction between the two modules.

Our actual work addresses more the problems inside each module. The
attachment module has been implemented at 99\%. Presently we are
working on the extension of the anaphora module particularly to deal
also with the anaphoric definite noun phrases.

\section*{References}

Azzam, S. 1994. CLAM COBALT conceptual analyser (COBALT Tech. Report
Del6.2). CRIL Ingénierie.

Carter, D. 1987.  Interpreting Anaphors in natural language Texts. Chichester:
Ellis Horwood.

Frazier, L. and Fodor, J. 1979.  The sausage machine: A New Two-Stage
Parsing Model, Cognition, 6.

Hobbs, J.R., and Bear, J. 1990. Two Principles of Parse Reference in
Proceedings of the 13th International
Conference on Computational Linguistics - COLING/90, vol. 3, Karlgren,
H., ed.  Helsinki:  University  Press.

Reinhart, T. 1983. Anaphora and Semantic Interpretation. London : Croom Helm.

Sidner, C.L. 1983. Focusing for Interpretation of pronouns. American Journal
of Computational Linguistics, 7,  217-231.

Wilks, Y., Huang, X., and Fass, D. 1985.  Syntax, Preference and Right
Attachment,  IJCAI.

Zarri, G.P. 1992. The descriptive component of hybrid knowledge
representation language, In: Semantic networks in Artificial
Intelligence, Lehmann, F., ed. Oxford: Pergamon Press.

\end{document}